\documentclass{article}
\usepackage{spconf,amsmath,graphicx}

\usepackage{enumitem}
\usepackage{caption}
\usepackage{comment}
\usepackage{xcolor}
\usepackage{amsfonts}
\usepackage{ragged2e}
 \usepackage{amsmath}
\usepackage{multirow}
\usepackage{booktabs}

\newcommand{\egno}{\textit{e}.\textit{g}.} 

\title{REGION-ADAPTIVE VIDEO SHARPENING VIA RATE-PERCEPTION OPTIMIZATION}

\name{Yingxue Pang, Shijie Zhao\sthanks{Corresponding author}, Mengxi Guo, Junlin Li, Li Zhang}
\address{Bytedance Inc.}

\begin{document}
%
\maketitle
\begin{abstract}
Sharpening is a widely adopted video enhancement technique. However, uniform sharpening intensity ignores texture variations, degrading video quality. Sharpening also increases bitrate, and there's a lack of techniques to optimally allocate these additional bits across diverse regions. Thus, this paper proposes RPO-AdaSharp, an end-to-end region-adaptive video sharpening model for both perceptual enhancement and bitrate savings. We use the coding tree unit (CTU) partition mask as prior information to guide and constrain the allocation of increased bits. Experiments on benchmarks demonstrate the effectiveness of the proposed model qualitatively and quantitatively.
\end{abstract}

\section{Introduction}
\label{sec:intro}
Sharpening~\cite{USM,USM2}, particularly with the use of the Unsharp Masking (USM) algorithm, is a widely adopted and dependable video enhancement technique. It reinforces high-frequency components, thereby enhancing edge and detail information in the video, resulting in a significant visual quality improvement. In detail, the USM algorithm involves generating a blurred version $B(x,y)$ of the original image $I(x,y)$, subtracting this from the original to extract high-frequency components, and subsequently amplifying and re-adding the high-frequency information to improve sharpness and clarity of the video. Mathematically, the sharpened result $I_{usm}(x,y)$ can be acquired by:
\begin{equation}
    I_{usm}(x,y) = I(x,y) + \alpha (I(x,y) - B(x,y)).
    \label{eq:usm}
\end{equation}

\begin{figure}[!t]
\centering
  \includegraphics[width=0.45\textwidth]{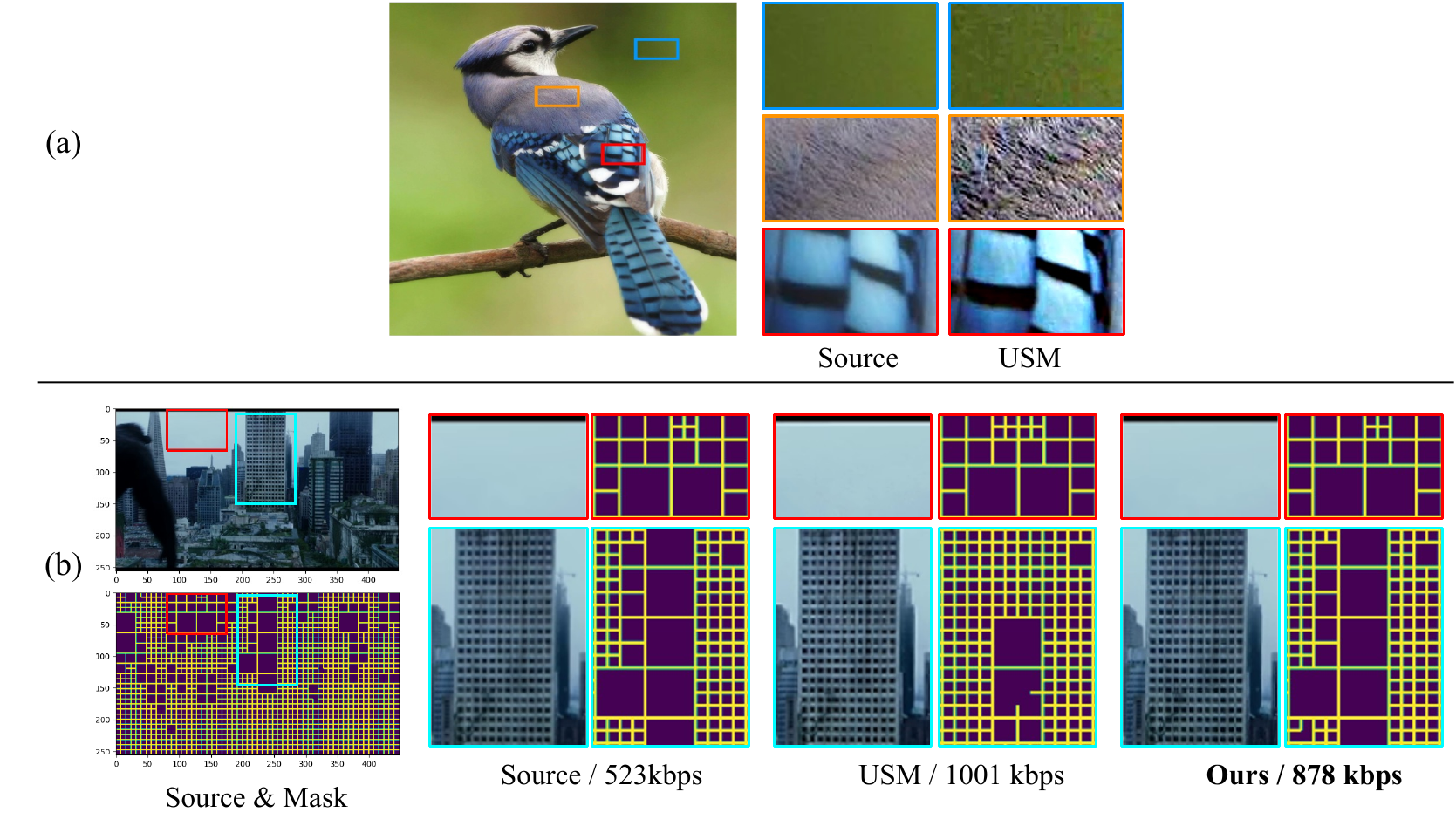}
  \caption{The illustration of our motivation.}
  \label{fig:motivation}
  \vspace{-5mm}
\end{figure}

However, uniformly applying a high-frequency amplification factor $\alpha$ to all regions of one frame is inappropriate due to texture variations, which can degrade image quality~\cite{sharpeen1, sharpeen2, sharpeen3}. As shown in Fig.~\ref{fig:motivation} (a), the same sharpening intensity applied to different regions yields varying subjective quality. In particular, blocks taken from areas with flat backgrounds experience a terrible amplification of previously negligible noise, while those from densely feathered areas exhibit overshoot artifacts. Conversely, moderately textured regions produce visually improved results after sharpening, as shown in the red box. Moreover, sharpening involves overlaying high-frequency information onto an image, resulting in increased video bitrate. Current algorithms lack consideration for the allocation of these added bits.

Therefore, we propose region-adaptive video sharpening by optimizing both subjective perception and bitrate savings. We find that the coding tree unit (CTU) partition~\cite{hevc, yang2019low, ccetinkaya2021ctu} provides direct insight into the impact of different regions on both bitrate and video quality. Using it as prior information to guide region-adaptive sharpening, the allocation of increased bits across regions after sharpening is naturally constrained. Specifically, we first preprocess different regions with varying levels of degradation using the CTU partition mask to obtain training pairs in Section~\ref{sec:dataprocess}. Second, we propose \textbf{RPO-AdaSharp}, a \textbf{R}ate-\textbf{P}erception \textbf{O}ptimized region-\textbf{Ada}ptive video \textbf{Sharp}ening model. It integrates CPEnhancer and neural video codec (NVC) into an end-to-end training framework, jointly optimizing perceptual enhancement and efficient bitrate savings. As shown in Fig.~\ref{fig:motivation} (b), we visualize one frame and corresponding CTU partition mask of the original video compression, the USM-enhanced compression, and our RPO-AdaSharp-enhanced compression. The mask of USM divides larger coding unit (CU) blocks into smaller CU blocks which require more bits. It results in unnecessary bitrate increases due to misplaced identification of noise and artifacts as important coding features. 
Our RPO-AdaSharp based on rate-perception optimization achieves a more similar CTU partition result or bits allocation to the original video, indicating a reasonable allocation of increased bits to regions that significantly influence video quality.

\section{Degradation Preprocessing}
\label{sec:dataprocess}
Using rate-distortion optimization (RDO)~\cite{hevc}, the optimal CTU partition is determined to minimize CTU quality distortion while restricting total bits. Larger CU sizes reduce bits and increase coding efficiency, while smaller CU sizes require more bits and have a greater impact on picture quality. Thus, we propose to utilize partitioning results as prior information to guide region-adaptive sharpening and constrain the allocation of additional bits. However, directly applying diverse sharpening intensities to CU blocks solely based on the CTU partition mask can result in terrible blocking artifacts arising from discontinuities in CU block sizes. To address this issue and construct training pairs, we counter the USM sharpening process in Eq.~\ref{eq:usm} by assuming the high-quality ground truth as the region-adaptive sharpening output $GT=I_{usm}(x,y)$ and incorporating distinct degrees of Gaussian blur to varied regions based on the CTU partition mask to create low-quality input data $LQ=I(x,y)$. Mathematically,
\begin{equation}
    LQ = \frac{GT + \alpha B(x,y)}{1 + \alpha}.
    \label{eq:usm_blur}
\end{equation}
where the high-frequency amplification factor $\alpha$ in Eq.~\ref{eq:usm} is utilized to control the degree of blurring here. 

\begin{figure}[!t]
\centering
\includegraphics[width=0.45\textwidth]{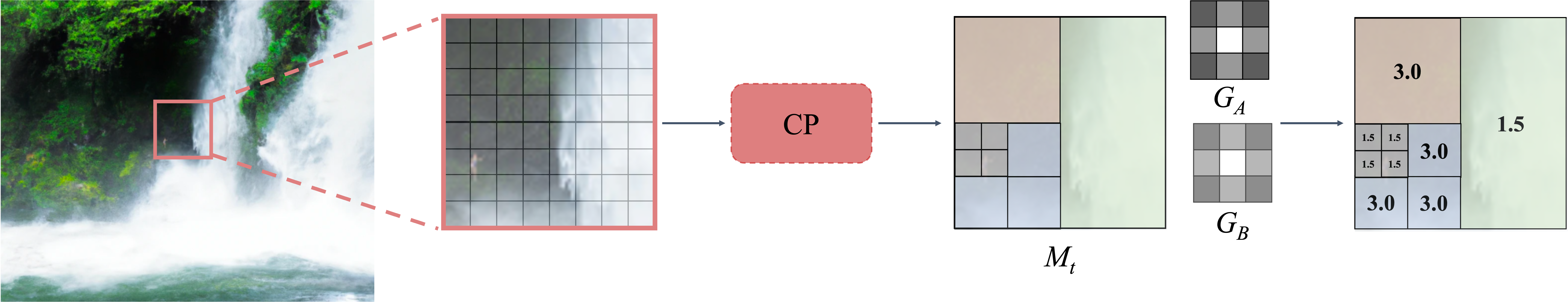}
\caption{Preprocessing of data degradation using the CTU partition mask.}
\label{fig:data}
\vspace{-5mm}
\end{figure}

Specifically, Fig.~\ref{fig:data} illustrates the entire degradation process, beginning with obtaining the CTU partition mask of the input based on the CP module. Subsequently, different degrees of Gaussian blurring with a kernel size of $3\times3$ are applied to distinct CU sizes. Since amplification of noise and overshoot artifacts commonly occur in texture-flat and rich regions, resulting in unnecessary increases in bitrate, the degradation degree is set to be lower (\egno, 1.5) for CU sizes of $8\times8$ and $64\times64$, while being higher (\egno, 3.0) for others.

\begin{figure*}[!t]
\centering
\includegraphics[width=\textwidth]{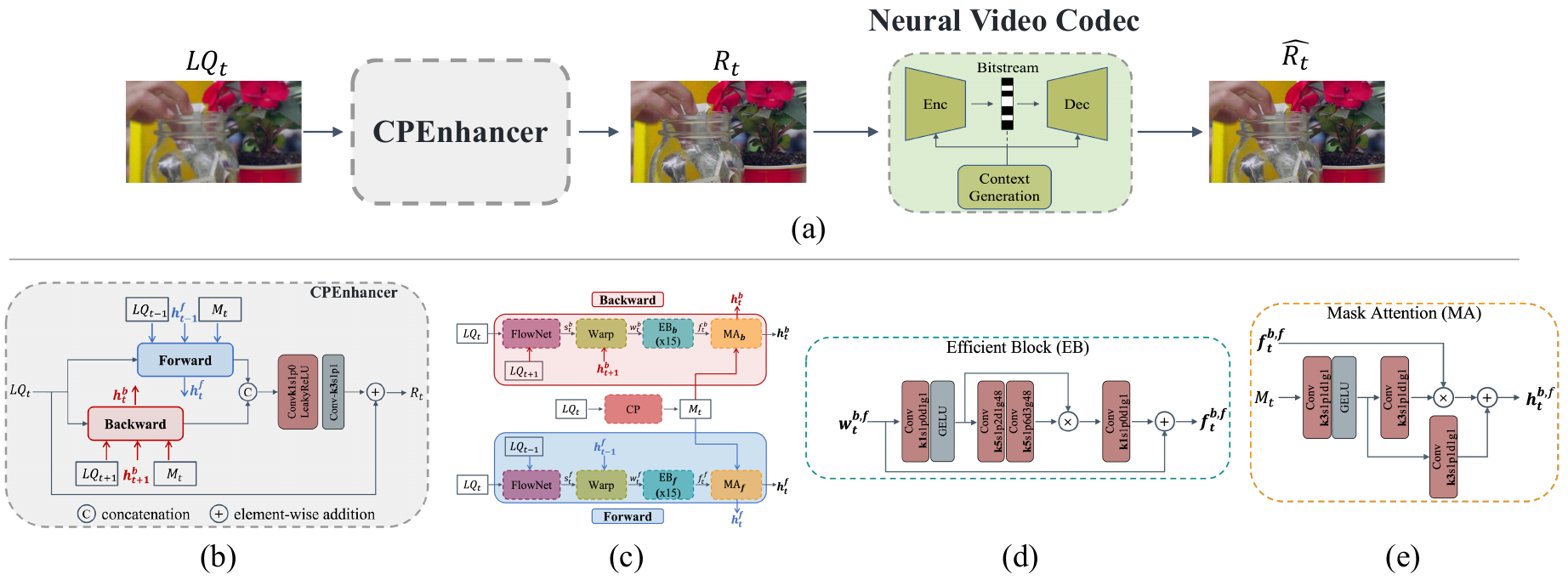}
\caption{Overall framework for our proposed RPO-AdaSharp. (\textbf{Zoom-in for best view})}
\label{fig:framework}
\vspace{-5mm}
\end{figure*}

\section{RPO-AdaSharp}
\label{sec:model}
To optimize perceptual enhancement while considering bitrate savings, we further propose RPO-AdaSharp, a Rate-Perception Optimized region-Adaptive video Sharpening model trained in an end-to-end fashion. As shown in Fig.~\ref{fig:framework} (a), our RPO-AdaSharp consists of two components: a CTU partition mask-based video enhancer (\textbf{CPEnhancer}) that minimizes perceptual reconstruction loss to enhance subjective quality, and a neural video codec (\textbf{NVC}) that minimizes rate-distortion (RD) loss to reduce the bits required to maintain the quality of the enhanced input. In detail, our CPEnhancer learns to enhance a low-quality input frame $LQ_{t}$ into an output frame $R_{t}$. The sharpened and enhanced frame $R_{t}$ is subsequently compressed and reconstructed into the frame $\hat{R}_{t}$ with NVC, which learns to use the least bits to get the best reconstruction quality.

\subsection{CPEnhancer}
As shown in Fig.~\ref{fig:framework} (b), to effectively capture long-term and global contextual information of the input video sequence, our CPEnhancer utilizes bidirectional propagation to independently propagate spatiotemporal features forward and backward in time similar to previous works~\cite{chan2021basicvsr,chan2022basicvsr++}. Given the low-quality sequence $LQ_{t}$, its CTU partition mask $M_{t}$ generated from the CTU partitioning (CP) module, its neighboring frames $LQ_{t-1}$ and $LQ_{t+1}$, and the corresponding features propagated from its neighbors $h_{t+1}^{b}$ and $h_{t-1}^{f}$, we get the propagated features $h_{t}^{b,f}$:
\begin{equation}
    \begin{aligned}
    & M_{t} = CP(LQ_{t}),\\
    & h_{t}^{b} = B_{b}(LQ_{t}, LQ_{t+1}, M_{t}, h_{t+1}^{b}), \\
    & h_{t}^{f} = F_{f}(LQ_{t}, LQ_{t-1}, M_{t}, h_{t-1}^{f}).
    \end{aligned}
\end{equation}
where $B_{b}$ and $F_{f}$ denote the backward and forward propagation branches respectively. 
Then the two propagated features are simply concatenated (Concat) and aggregated with a two-layer convolution (Conv). Here we denote each convolutional layer with the corresponding kernel size (k), stride (s), and padding size (p). Finally, the input sequence is added back to generate the final output sequence $R_{t}$:
\begin{equation}
    R_{t} = Conv(Concat(h_{t}^{b}, h_{t}^{f})) + LQ_{t}
\end{equation}

The architecture of the bidirectional propagation branches, $B_{b}$ and $F_{f}$, is depicted in Fig.~\ref{fig:framework} (c), which comprises flow estimation network (FlowNet), spatial warping (Warp), efficient blocks (EB), CTU partitioning (CP), and mask attention (MA). FlowNet is responsible for estimating flow features $s_{t}^{b,f}$, while the Warp module warps feature to achieve spatial alignment $w_{t}^{b,f}$,
\begin{equation}
\begin{aligned}
      & s_{t}^{b,f} = FlowNet(LQ_{t}, LQ_{t\pm1}), \\
    & w_{t}^{b,f} = Warp(s_{t}^{b,f}, h_{t\pm1}^{b,f}).
\end{aligned}
\end{equation}
The aligned features are then inputted into a stack of efficient blocks (EB) in Fig.~\ref{fig:framework} (d) for feature extraction and refinement $f_{t}^{b,f}$,
\begin{equation}
    f_{t}^{b,f} = EB_{b,f}(w_{t}^{b,f}).
\end{equation}
Lastly, a mask attention (MA) module in Fig.~\ref{fig:framework} (e) is integrated, which employs spatially adaptive, learned affine transformations to modulate the extracted features $f_{t}^{b,f}$ through CTU partition mask $M_{t}$ generated from the CP module,
\begin{equation}
    h_{t}^{b,f} = MA_{b,f}(M_{t}, f_{t}^{b,f}).
\end{equation}
Utilizing the MA, mask information propagates throughout the network to emphasize the significance of local masked feature regions through spatially varying transformation. Consequently, region-adaptive enhancement is achieved effectively.

\subsection{NVC}
Our RPO-AdaSharp approach aims to reduce the bitrate increase while maintaining the enhanced subjective effect of the sharpened output after compression. We explore two strategies for bitrate saving. Firstly, we introduce CTU partition masks for region-adaptive sharpening, aiming to allocate increased bits to regions that would benefit the most from improved visual quality. Secondly, we suggest incorporating a video codec to globally control the bits allocation of the enhancement results. The codec can measure quality distortion and bitrate of the sharpening output after compression, maintaining the enhanced quality and reducing the bitrate through rate-distortion optimization. However, standard video codecs are non-differentiable and blocky, making them unsuitable for direct incorporation into our end-to-end optimization framework. Thus, we adopt a Neural Video Codec (NVC) from~\cite{dcvc-dc} to approximate the rate-distortion behavior of standard codecs. As illustrated in Fig.~\ref{fig:framework} (a), we feed the sharpened output $R_{t}$ into the NVC for end-to-end learning of the rate-distortion loss, thereby preserving the enhanced quality or reducing the compression distortion while using the least bits.

\subsection{Objective Functions}
We adopt the robust Charbonnier loss~\cite{charbonnier1994two,lai2018fast} as our perceptual reconstruction loss for sharpening, 
\begin{equation}
    \mathcal{L}_{rec} = \sqrt{||GT_{t}-R_{t}||^{2} + \epsilon^{2}}.
\end{equation}
where $\epsilon=1\times10^{-12}$ is a fixed value.
Completely referring to ~\cite{dcvc-dc}, the bitrate cost $\mathcal{R}$ represents the bits used for encoding the quantized latent representation of bitstream and motion vector, both associated with the bits used for encoding their corresponding hyper prior.
And we define the compression distortion $\mathcal{D}$ to measure the error between reconstructed frame $\hat{R}_{t}$ and ground truth $GT_{t}$ using L2 distance. The rate-distortion loss is written as:
\begin{equation}
    \mathcal{L}_{rd} = \mathcal{R} + \lambda \cdot \mathcal{D}(GT_{t}, \hat{R}_{t}).
\end{equation}
where $\lambda$ controls the tradeoff between the distortion $\mathcal{D}$ and the bitrate cost $\mathcal{R}$. 
The overall loss function as the weighted sum of perceptual reconstruction loss $\mathcal{L}_{rec}$ and rate-distortion loss $\mathcal{L}_{rd}$ as follows:
\begin{equation} 
\label{eq:overall}
    \mathcal{L}_{overall} = \mathcal{L}_{rec} + \gamma \mathcal{L}_{rd}.
\end{equation}

\section{Experiments}
\label{sec:exp}
\subsection{Implementation Details}
The training dataset is Vimeo90k~\cite{xue2019video}. We first pretrain CPEnhancer for 100k iterations with $\mathcal{L}_{rec}$ loss using initial learning rates of $1\times 10^{-4}$. Then we finetune CPEnhancer for 100k iterations while keeping the weights of NVC fixed using $\mathcal{L}_{overall}$ loss defined in Eq.~\ref{eq:overall} with $\gamma=10$ and $\lambda=85$. We adjust the learning rates of CPEnhancer to $5\times 10^{-5}$. We employ the Adam and the mini-batch size is set as 8. We train with a sequence of 7 frames. The GOP size of NVC is set to 4. We use pretrained SPyNet~\cite{flownet}, DPNet~\cite{CTUmusk}, and DCVC-DC~\cite{dcvc-dc} as our FlowNet, CP, and NVC respectively. Our model is implemented based on the PyTorch framework with 4 NVIDIA A100-SXM-80GB GPUs.

In terms of testing, we randomly sample 50 videos from two datasets respectively: KoNViD-1k~\cite{hosu2017konstanz} and LIVE-VQC~\cite{sinno2018large}. The NVC is replaced with a standard video codec to perform real RD behaviors. 
For standard codecs, we employ the constant rate factor (CRF) setting for H.264 and H.265, and the quantization parameter (QP) setting for H.266. The CRF/QP ranges used are [21, 24, 27, 30, 33], and the encoding preset is set to \textit{medium}.

\begin{table}
\centering
\caption{BD-Rate comparison on VMAF~\cite{vmaf}. The anchor is to compress videos with H.264, H.265, or H.266 without sharpening.}
\label{table:cmp}
\resizebox{0.8\linewidth}{!}{
\begin{tabular}{ccc}
\toprule
                      & KoNViD-1k (50)    & LIVE-VQC (50) \\ \midrule
usm 1.5 + H.264       & 16.88\%           & 22.00\%                           \\
usm 3.0 + H.264       & 42.25\%           & 78.19\%                           \\
\textbf{Ours + H.264} & \textbf{-11.08\%} & \textbf{-18.15\%}                 \\\hline
usm 1.5 + H.265       & 31.85\%           & 29.01\%                           \\
usm 3.0 + H.265       & 69.03\%           & 113.56\%                          \\
\textbf{Ours + H.265} & \textbf{-10.27\%} & \textbf{-20.66\%}                 \\ \hline
usm 1.5 + H.266       & 50.47\%           & 73.65\%                           \\
usm 3.0 + H.266       & 123.13\%          & 245.73\%                          \\
\textbf{Ours + H.266} & \textbf{-13.55\%} & \textbf{-24.64\%}  \\
\bottomrule
\vspace{-4mm}
\end{tabular}}
\end{table}

\begin{table}
\centering
\caption{Perceptual quality comparison on KoNViD-1k(50)/LIVE-VQC (50) with various metrics at a fixed bit rate of 800k.}
\label{table:cmp_2}
\resizebox{\linewidth}{!}{
\begin{tabular}{ccccc}
\toprule
method       & VMAF($\uparrow$)~\cite{vmaf}   & MS-SSIM($\uparrow$)~\cite{msssim}   & CLIP-IQA($\uparrow$)~\cite{clipiqa}  & NIQE($\downarrow$)~\cite{niqe}   \\ 
\midrule
USM 1.5 + H.265       & 89.995/83.143  & 0.966/0.946  & 0.256/0.267  & 6.083/4.926                \\
USM 3.0 + H.265       & 89.036/81.549  & 0.945/0.910  & 0.255/0.263  & 6.396/4.815                \\
H.265                 & 90.932/84.277  & 0.977/0.961  & 0.242/0.246  & 6.742/5.233                \\
\textbf{Ours + H.265} & \textbf{91.745}/\textbf{85.449} & \textbf{0.981}/\textbf{0.969} & \textbf{0.263}            /\textbf{0.270}       & \textbf{5.703}/\textbf{4.734}       \\ 
\bottomrule
\end{tabular}}
\end{table}

\subsection{Experimental Results}
As the data degradation is based on $\alpha=1.5$ and $\alpha=3.0$, we conduct performance comparisons against USM with sharpening levels 1.5 (USM 1.5) and 3.0 (USM 3.0).

\begin{figure}[!t]
\centering
\includegraphics[width=0.45\textwidth]{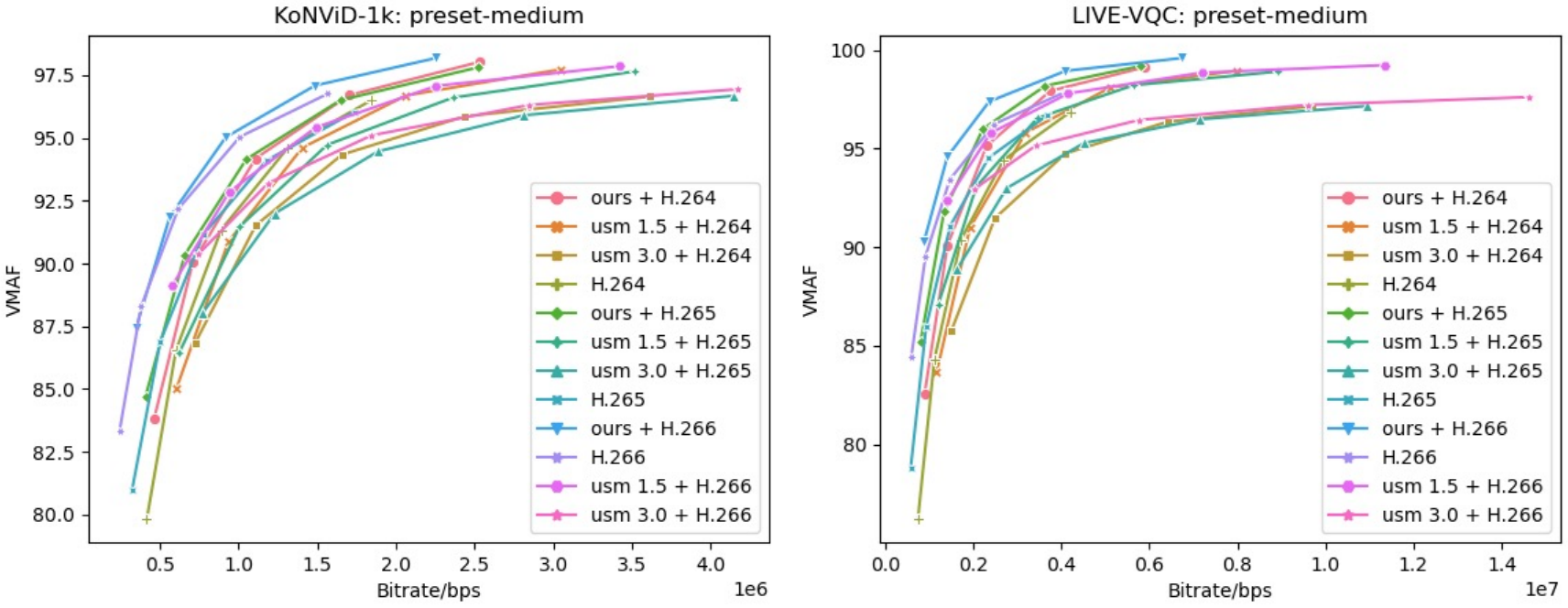}
\caption{RD curves on VMAF~\cite{vmaf}.}
\label{fig:cmp}
\vspace{-2mm}
\end{figure}

\begin{figure}[!t]
\centering
\includegraphics[width=0.45\textwidth]{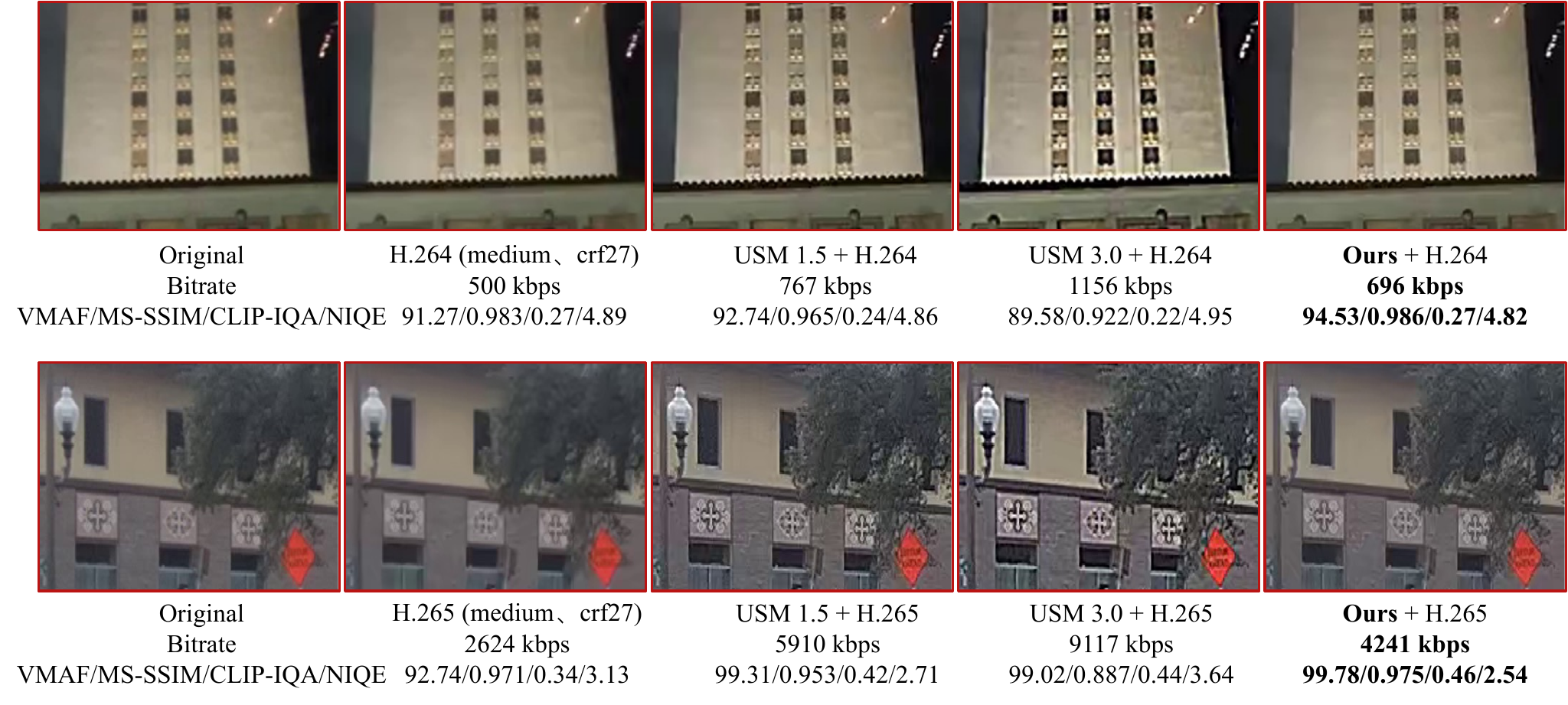}
\caption{The subjective quality comparison of enhancement after compression.(\textbf{Zoom-in for best view})}
\label{fig:vis}
\end{figure}

\begin{figure}[!t]
\centering
\includegraphics[width=0.45\textwidth]{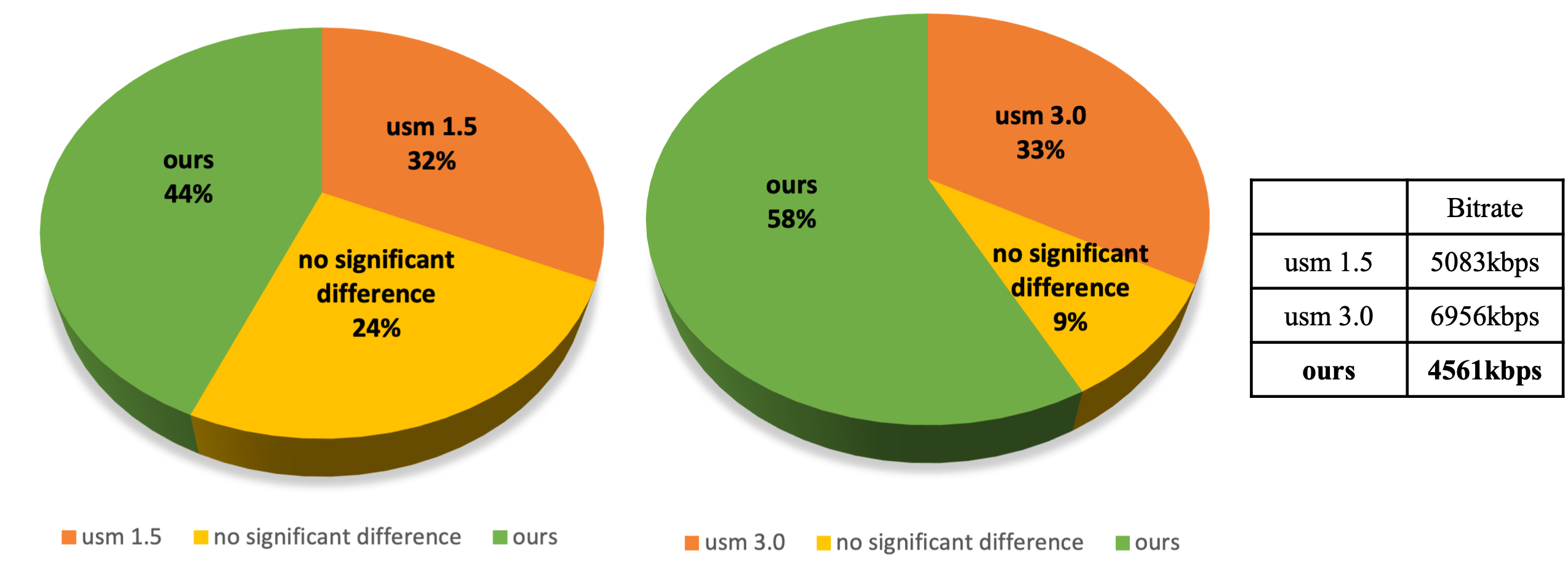}
\caption{User study results.}
\label{fig:user}
\vspace{-3mm}
\end{figure}

Table~\ref{table:cmp} shows the BD-Rate comparison using VMAF. On average, our sharpening frames require $14.62\%$, $15.46\%$ and $19.09\%$ fewer bits than the anchor to achieve comparable video quality. In contrast, USM 1.5 + codec and USM 3.0 + codec perform dramatically poorly compared to the anchor and require much more bits for encoding. In Fig.~\ref{fig:cmp}, we can see that our RPO-AdaSharp + codec substantially outperforms the anchors (H.264, H.265, H.266), USM 1.5 + codec and USM 3.0 + codec.

We quantitatively compare visual performance at a fixed bitrate of 800k in Table~\ref{table:cmp_2}. The results show that Ours+H.265 achieves superior perception scores at a comparable bitrate, indicating significant BD-Rate gains. Fig.~\ref{fig:vis} illustrates qualitative visual comparisons with default testing encoding settings, revealing reduced noise and overshoot artifacts in our processed frames compared to USM 1.5 and USM 3.0 after compression, with a minimal increase in bitrate. Overall, our RPO-AdaSharp proves effective in optimizing visual quality enhancement while achieving bitrate savings.

We further conduct a user study to compare the subjective effects of our method and USM. We randomly select 10 videos from the testing data. Each video is displayed with three results on a webpage: A (H.265), B (our method + H.265), and C (USM + H.265). Participants ($n=20$) are asked to select their preferred result between B and C (in random order), or indicate no significant difference. We collect a total of 400 responses, with 200 responses obtained for each comparison (10 videos multiplied by 20 participants). Fig.~\ref{fig:user} shows the user study results, indicating that our method consistently received higher preference.

\section{Conclusion}
In this paper, we propose a region-adaptive video sharpening approach that simultaneously optimizes perception improvement and bitrate saving. Our method first employs the CTU partition mask to degrade the ground truth by applying varying degrees of Gaussian blurring to different CU sizes, implicitly constraining the allocation of increased bits. We then introduce RPO-AdaSharp which integrates CPEnhancer and NVC to jointly enhance the perceptual quality while reducing bitrate requirements. Extensive experimental results demonstrate the effectiveness of our proposed method.




\vfill\pagebreak

\bibliographystyle{IEEEbib}
\bibliography{refs}

\end{document}